\documentclass{IEEEtran}

\usepackage{times}
\usepackage{graphicx}
\usepackage{subcaption}

\begin{document}

\title{Building Open Knowledge Graph for Metal-Organic Frameworks (MOF-KG): Challenges and Case Studies}

\author{Yuan An, Jane Greenberg, Xintong Zhao, Xiaohua Hu, Scott McCLellan, Alex Kalinowski\\
		\textsf{College of Computing and Informatics} \\
		\emph{Drexel University} \\
		\emph{Philadelphia, PA, USA} \\
		\texttt{Contact:\{ya45,jg3243,xh29\}@drexel.edu}  \\
		\vspace{5mm}
	\and
	Fernando J. Uribe-Romo, Kyle Langlois, Jacob Furst \\
	\textsf{Department of Chemistry} \\
	\emph{University of Central Florida} \\
	\emph{Orlando, FL, USA} \\
	\texttt{Contact:fernando@ucf.edu} \\
	\vspace{5mm}
	\and
	Diego A. Gómez-Gualdrón, Fernando Fajardo-Rojas, Katherine Ardila \\
	\textsf{Department of Chemical and Biological Engineering} \\
	\emph{Colorado School of Mines} \\
	\emph{Golden, CO, USA}\\
	\texttt{Contact:dgomezgualdron@mines.edu}
}

\maketitle
\thispagestyle{empty}

\begin{abstract}
  Metal-Organic Frameworks (MOFs) are a class of modular, porous crystalline materials 
  that have great potential to revolutionize applications such as gas storage, molecular 
  separations, chemical sensing, catalysis, and drug delivery. The Cambridge Structural 
  Database (CSD) reports 10,636 synthesized MOF crystals which in addition contains  
  ca. 114,373 MOF-like structures.  
  The sheer number of synthesized (plus potentially synthesizable) 
  MOF structures requires researchers pursue computational techniques to screen and 
  isolate MOF candidates. In this demo paper, we describe our effort on leveraging 
  knowledge graph methods to facilitate MOF prediction, discovery, and synthesis.  We 
  present challenges and case studies about (1) construction of a MOF knowledge graph 
  (MOF-KG) from structured and unstructured sources and (2) leveraging the MOF-KG for 
  discovery of new or missing knowledge. 
\end{abstract}

\pagestyle{empty}

\section{INTRODUCTION}
\label{sec:introduction}

\begin{figure}[t]
	\centering
	\includegraphics[width=.45\textwidth]{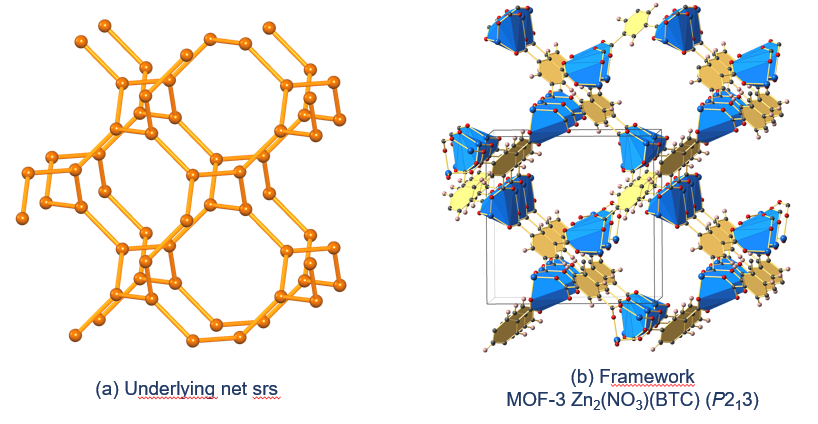}
	\caption{The Underlying Net and Framework of MOF-3}
	\label{fig:underlying-net-MOF-3}
\end{figure}

Metal-Organic Frameworks (MOFs) are materials that possess modular, porous 
crystal structures that can be (conceptually) modified by “swapping” constituent 
building blocks. MOF building blocks correspond to metal-based clusters and organic 
linkers, which are interconnected in patterns described by an underlying net (Figure \ref{fig:underlying-net-MOF-3}). 
MOFs have great potential to revolutionize applications such as gas storage, molecular
separations, sensing, catalysis, and drug delivery, primarily due to their usually high 
surface area and exceptionally tunable properties \cite{ref1}. 
But the combinatorics of building blocks means that chemists have access to a 
(not fully explored) “material design space” of trillions of structures. To date, 
there are 10,636 synthesized MOF crystals reported in the Cambridge Structural Database 
(CSD) \cite{ref2} which in addition contains ca. 114,373 
MOF-like structures.  The sheer number has made the identification of optimal MOFs 
(and subsequent) synthesis for a given application a very challenging task. 
Thus, considerable efforts have been put into developing effective computational 
techniques to screen and isolate candidate MOF structures for the application of choice.  

Previous efforts include the creation of large MOF databases  and development of high 
throughput automated workflows such as molecular simulation and machine learning for 
MOF property prediction. MOF databases contains both synthesized and “hypothesized” 
MOF structures \cite{ref3}. When a hypothesized MOF structure is identified as 
promising, it is unclear if and/or how such structure can be realized synthetically.  
This is partly due to the crystallization process leading to MOF formation not being 
fully understood. Worse, a large amount of MOF synthesis data are not readily available 
for computation but scattered in scientific literature. 
The separation between the crystals’ structure information and their synthesis data 
exacerbates the difficulty of screening MOFs.

We have undertaken a project aiming to leverage advanced knowledge graph methods 
to facilitate MOF prediction, discovery, and synthesis.  In this demo paper, we 
describe the challenges for building such an open knowledge graph for MOFs from 
structured and unstructured data. We present case studies on addressing several challenges.

Knowledge graph (KG) represents the knowledge in a domain using a graphical structure 
consisting of (typed-) vertices and (typed-) links.  Despite the growing number of 
materials science related databases \cite{ref4}, knowledge graphs built for materials 
science domains are still rare \cite{ref5,ref6}. There is no applicable knowledge graph 
for MOFs or Reticular Chemistry in general. The \textbf{\emph{first challenge}} we face in 
building such a knowledge graph is to choose the underlying graph architecture, 
for example, whether using RDF triple store or labeled property graph model. 
Materials science data typically describe numeric electronic, chemical, and 
physical properties. These properties may connect through different relationships 
for guiding the design of multifunctional materials. Additional properties 
can be extracted from the literature. Considering the need of capturing the 
properties, we choose the labeled 
property graph model for the underlying architecture and implement 
it in the Neo4j platform.   

\begin{figure}[!ht]
	\centering
	\includegraphics[width=.45\textwidth]{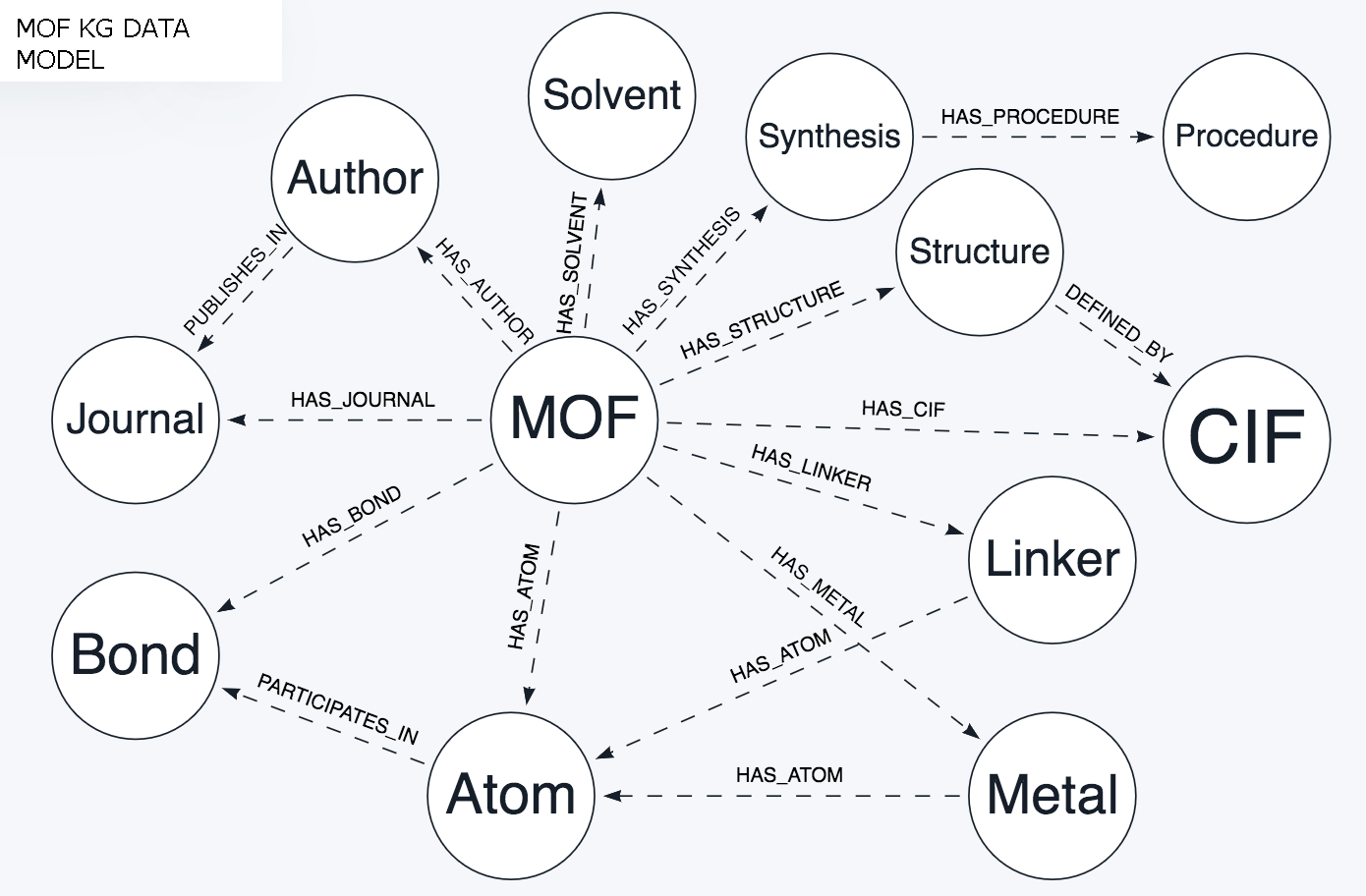}
	\caption{Partial Backbone of the Data Model for the MOF-KG.}
	\label{fig:MOF_KG_data_model}
\end{figure}

\section{DEFINING A DATA MODEL FOR THE MOF KNOWLEDGE GRAPH (MOF-KG)}
\label{sec:data-model}

The knowledge graph for MOFs (MOF-KG) should facilitate researchers in reticular 
chemistry to practice their distinguished activities. The past decades have seen 
an explosive increase in synthesis and characterization of MOFs. A systematic workflow 
pattern in the field has emerged \cite{ref7}.  The pattern is an iterative cycle consisting of 
3 refinement steps: \emph{Synthesis, Activation}, and \emph{Analysis}.  The workflow typically 
starts with a \emph{Synthesis} step where the researchers will screen and identify various 
crystalline structures by considering many parameters including  starting compounds, 
molar ratios, temperature, concentration, reaction additives, modulators, solvents, 
and reaction time. The researchers can apply computational techniques such as 
high-throughput synthesis or screening to aid their experimental process \cite{ref8}.

Next at \emph{Activation}, the researchers will assess the permanent porosity and architectural 
stability of the crystal by removing all guest molecules (including solvent) from the 
pores of the framework without causing collapse of its structure. The third step is 
\emph{Analysis} where the researchers will identify and characterize the physical and chemical 
properties of the crystal. Here, the researchers may resort to analytical, 
computational, and machine learning tools to study and characterize porosity \cite{ref9}, 
gas uptake \cite{ref10}, stability \cite{ref11}, and other important properties \cite{ref12}. 
The workflow will cycle through these 3 steps iteratively until successful results are obtained.

Due to the rapid development in reticular chemistry,  there is no a general agreed 
system of nomenclature for describing MOFs and related activities. A couple of 
initiatives have worked on standardizing terminologies \cite{ref13,ref14}, although the diversity 
in the focus and the scientific inquiry has led to a variety of terminological usages 
for this class of compounds. The \textbf{\emph{second challenge}} in building the MOF-KG is to define a 
data model to conceptualize MOF and its related activities corresponding to the workflow pattern.

To address this challenge, we define a data model for the MOF-KG with 4 major areas: 
\emph{synthesis, structure, atomic composition}, and \emph{publication}. Figure \ref{fig:MOF_KG_data_model} illustrates a 
partial backbone 
of the data model with main concepts and relationships. Each concept and relationship has 
its own set of properties. The data model is continuously being refined and extended 
with newly identified concepts and additional relationships. 
\begin{figure}[!ht]
	\centering
	\includegraphics[width=.45\textwidth]{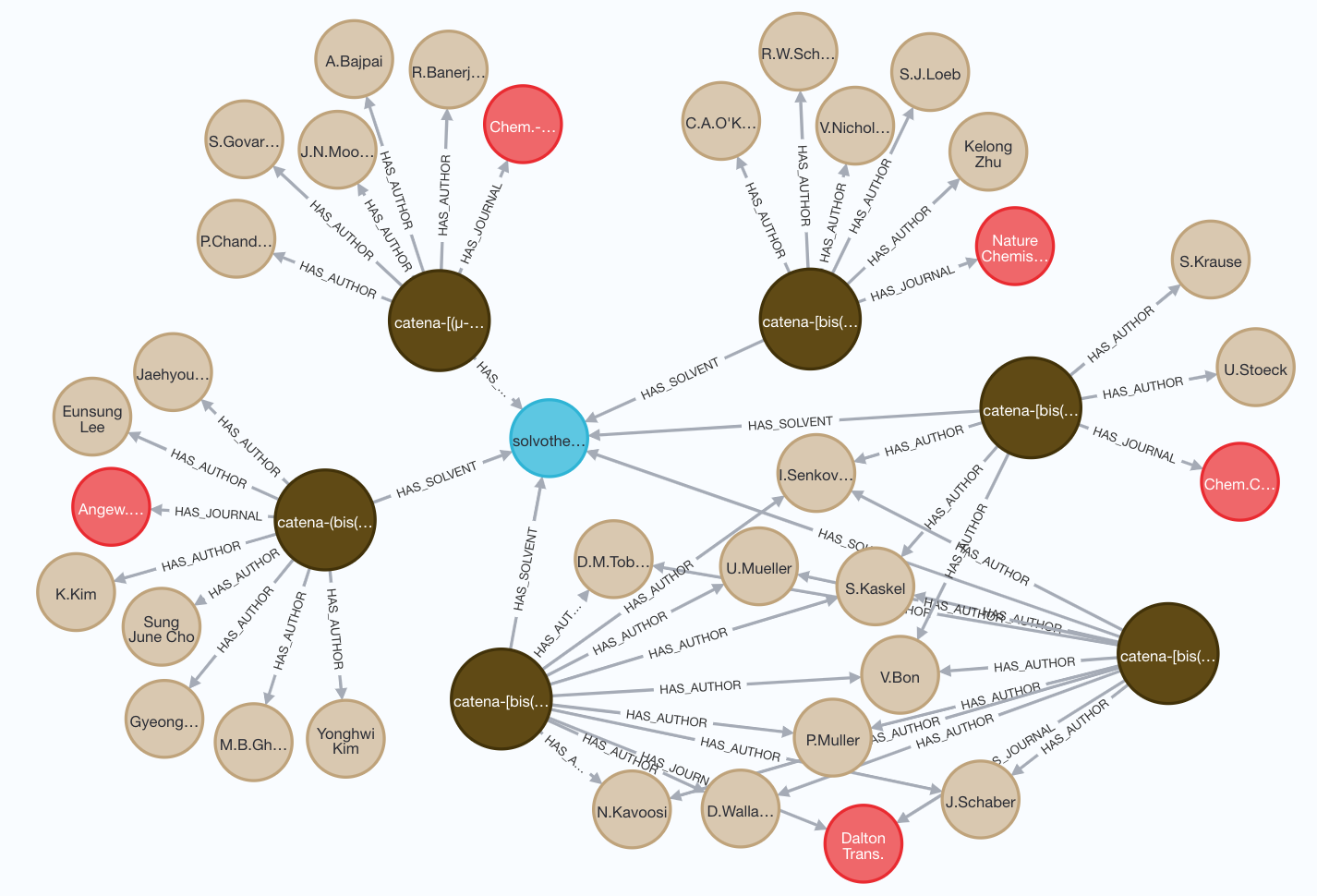}
	\caption{Excerpt of the MOF-KG Showing the Connections between Journals, Authors, the Same Solvent, and MOFs.}
	\label{fig:MOF_KG_mof_journal}
\end{figure}

\section{BUILDING THE MOF-KG FROM THE CSD MOF COLLECTION}
\label{sec:building-CSD-MOF}

\begin{figure*}[!th]
	\centering
	\subfloat[A published paragraph describing a synthesis procedure]{
		\includegraphics[width=0.45\linewidth]{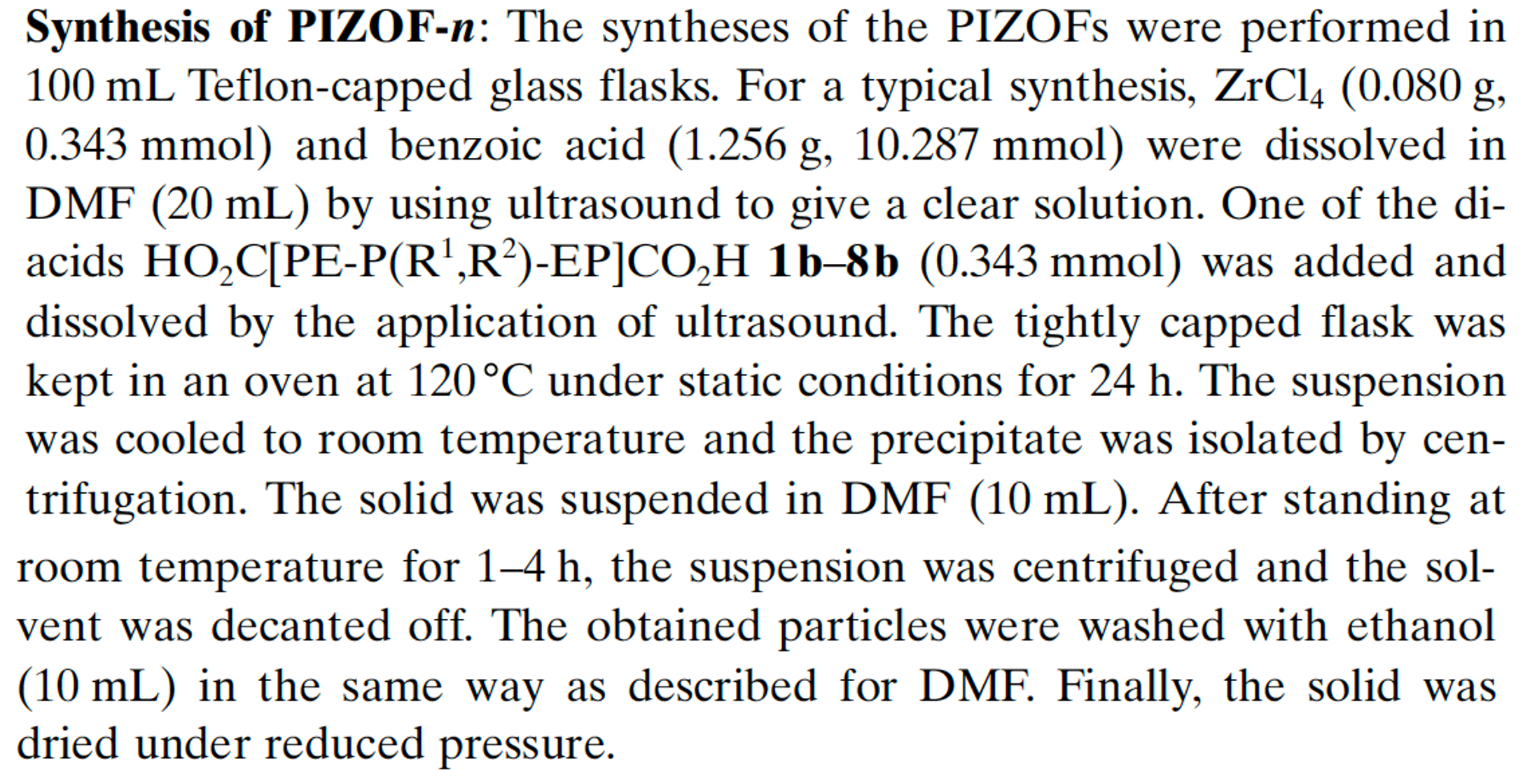} 
		\label{fig:synthesis-paragraph}
	}
	\subfloat[The extracted synthesis procedure steps from the paragraph on left]{
		\includegraphics[width=0.45\linewidth]{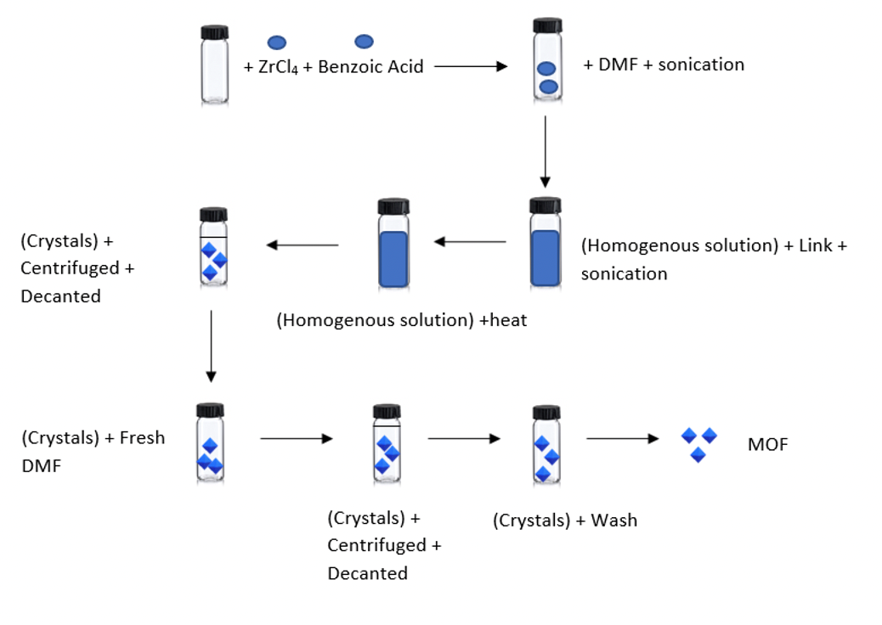}
		\label{fig:synthesis-procedure}
	}
	\caption{Extracting synthesis procedure steps from synthesis paragraph.
	}
	\label{fig:synthesis-paragraph-procedure}
\end{figure*}   
To instantiate the data model with instances for building the KG, we 
extract MOF related entities, relationships, and properties from both 
structured and unstructured sources. For structured sources, we currently 
focus on the Cambridge Structural Database (CSD)  curated by the Cambridge 
Crystallographic Data Centre (CCDC), a world-leading organization that 
compiles and maintains small-molecule organic and metal-organic crystal 
structures. An important feature of the CSD is that it contains successful 
structures (crystals) that have been realized in experiment, with crystal 
structure experimentally measured and solved using diffraction techniques 
(X-rays, neutrons, electrons). The database contains data such as crystal 
symmetry, atom positions, occupancy, and displacements in the form of a CIF  
(crystallographic information file) which is a highly standard format for 
storing crystallographic structural data and metadata. The \textbf{\emph{third challenge}} 
for us is to map the structured information in CSD to the MOF-KG described 
by the data model.  To address the challenge, we develop a schema mapping tool 
to (semi-)automatically extract, transform, and load data from CSD to the MOF-KG.

Once the MOF-KG is populated with the instances extracted from the CSD database, 
we can issue queries against the knowledge graph to explore connections among the 
MOF related entities. Figure \ref{fig:MOF_KG_mof_journal} shows an excerpt from the resultant graph after 
issuing the following query: \emph{``retrieve the authors and publication journals of 
the MOFs that have the same solvent.''} 

\section{AUGMENTING THE MOF-KG WITH SCHOLARLY ARTICLES}
\label{sec:augmenting-MOF-KG}

While the curated databases such as CSD contain significant amounts 
of information about crystal structures, scholarly articles that are 
rapidly growing in quantities contain rich knowledge resources of the 
synthesis procedures. However, computers cannot recognize the sequence of 
synthesis actions reported in plain text (Figure \ref{fig:synthesis-paragraph}) \cite{ref15}. 
Hence, the \textbf{\emph{fourth 
challenge}} is to extract synthesis conditions from unstructured 
text to augment the MOF-KG (Figure \ref{fig:synthesis-procedure}). There is seldom a standardized way for 
reporting synthesis conditions. Every researcher presents synthesis in different 
manners, e.g., in main article, implicit or explicit, in appendix or supplementary 
information file. Also, these conditions are often reported incomplete, as many 
researchers have pre-assumptions of their readership. This incomplete/assumed 
information may hinder immediate reproducibility of the synthesis.

Some effort has been put into the use of natural language processing (NLP) 
techniques \cite{ref16,ref17,ref18,ref19,ref20,ref21} for synthesis extraction. Existing techniques include 
rule-based (e.g., regular expression, pre-defined rules) and deep learning/machine 
learning based approaches \cite{ref20,ref21}. As a case study, we apply a recently developed 
rule-based NLP approach \cite{ref19} to extract the synthesis information from 114 articles 
that match 114 MOFs in the CSD Collection. There are 46 synthesis routes were 
recognized and they turned out to be accurate. However, only three solvent records 
were extracted. We manually examined ten synthesis procedures reported in articles. 
The results indicate that solvents are indeed reported in text but not recognized by 
the extraction algorithm. One possible reason is that solvent information is 
described in various contexts and different ways (e.g., water, N,N-dimethyl formamide, DMF).

Rule-based approaches are convenient to implement but brittle to the text context. 
They suffer from the low recall problem. Deep learning and general machine learning 
based approaches show better performance than rule-based approaches, but they suffer 
from the lack of annotated training corpus.  Currently, we are developing weakly-supervised 
information extraction algorithms to address this problem.

\section{USING THE MOF-KG FOR DISCOVERING NEW AND MISSING KNOWLEDGE}
\label{sec:discovering-missing-knowledge}

Knowledge graphs often suffer from incompleteness. For example, even 
the state-of-the-art NLP techniques cannot extract all available information 
from text. A flurry of research has been conducted on knowledge graph 
completion by predicting missing links \cite{ref22}. A notable approach is to learn 
knowledge graph representations, that is, low-dimensional embedding vectors 
for downstream classification and prediction \cite{ref23}. The MOF-KG is inevitably 
incomplete due to either incomplete information in the CSD database or 
incomplete extraction from text.  We aim to develop MOF-KG related link 
prediction techniques for discovering new and missing information. 

As a case study, we choose to predict the \texttt{`HAS\_SOLVENT'} links in the MOF-KG. 
Solvent information is an important variable for MOF synthesis and 97\% of the 
solvent information is missing in the CSD MOF collection.  We extract the publication, 
author, atom, and bond information for the 268 MOFs that have solvent information in the
CSD collection. We apply various knowledge graph embedding models including 
\emph{TransE, ConvE, ComplEx, DistMult} 
and \emph{SimplE} on the data set. We randomly sample 20\% as the test data and
predict the \texttt{`HAS\_SOLVENT'} relation for them. 
The results are reported in 
Table \ref{tab::predicting-missing-solvent}.  We use the following rank-based evaluation metrics:
\begin{itemize}
	\item MRR (Inverse Harmonic Mean Rank): higher is better, range $[0, 1]$.
	\item Hits@K (with K as one of $\{1, 5, 10\}$): higher is better, range $[0, 1]$.
	\item AMRI (Adjusted Arithmetic Mean Rank Index): higher is better,  range $[-1, 1]$. 
	AMRI=0 means the model is not better than random scoring.
\end{itemize}

\begin{table}[!ht]
	\begin{center}
		\begin{tabular}{|l|r|r|r|r|r|}
			\hline
			\textbf{KGE Model} & \textbf{MRR} & \textbf{AMRI} & \textbf{Hits@10} & \textbf{Hits@5} & \textbf{Hits@1} \\
			\hline
			TransE	& 0.15	& 0.97	& 0.38	& 0.25	& 0.04 \\
			\hline
			ConvE	& 0.12	& 0.94	& 0.21	& 0.18	& 0.07 \\
			\hline
			ComplEx	& 0	& -0.13	& 0	& 0	& 0 \\
			\hline
			DistMult	& \textbf{0.25}	& \textbf{0.99}	& \textbf{0.5}	& \textbf{0.42}	& \textbf{0.14} \\
			\hline
			SimplE	& 0	& 0.11	& 0	& 0	& 0 \\
			\hline
		\end{tabular}
	\end{center}
	\caption{Apply Various Knowledge Graph Embedding (KGE)  Models for Predicting Missing Solvent information for MOFs}
	\label{tab::predicting-missing-solvent}
\end{table}

The results show that the DistMult KG embedding model achieved the best result in 
terms of all the metrics. The AMRI is close to 1 and Hits@10 is 0.5. With the 
simple information of publication and atomic element, a KG embedding can 
successfully predict important missing information. Encouraged by these 
results, we plan to develop and apply more sophisticated link prediction 
approaches to the MOF-KG. 

\section{CONCLUSION AND NEXT STEPS}
\label{sec:conclusion}

The challenges presented here show that building a knowledge graph for MOFs 
requires a strong synergy among materials scientists, chemists, informaticians,
and data and computer scientists. The case studies demonstrate the potentials
of knowledge graphs for discovering information that is missing in the original 
distributed and heterogeneous sources. This project is a part of the effort 
undertaken by the NSF Institute for Data Driven Dynamical Design (ID4). The next 
steps include enriching the MOF-KG and building   user-friendly natural language 
and chatbot query interfaces for domain scientists  to conduct further knowledge 
graph-empowered materials discovery.

\section*{Acknowledgment}
The research reported on in this paper is supported, in part, by the U.S. National Science Foundation, 
Office of Advanced Cyberinfrastructure (OAC): Grant: 1940239 and 2118201.

\bibliographystyle{ACM-Reference-Format}

\end{document}